\pdfoutput=1
\documentclass[11pt]{article}

\usepackage[final]{acl}

\usepackage{times}
\usepackage{latexsym}

\usepackage[T1]{fontenc}
\usepackage[utf8]{inputenc}

\usepackage{microtype}

\usepackage{inconsolata}

\usepackage{graphicx}

\usepackage{tabularx}
\usepackage{booktabs}
\usepackage{subcaption}
\usepackage{multirow}
\usepackage{pifont}
\usepackage{enumitem}

\newcommand{\benchmark}{\textsc{LongFormFact}}

\usepackage{amsmath}
\DeclareMathOperator*{\aggd}{{\textit{AGG\textsubscript{D}}}}
\DeclareMathOperator*{\aggs}{{\textit{AGG\textsubscript{S}}}}

\title{On Positional Bias of Faithfulness for Long-form Summarization }

\author{David Wan$^1$\thanks{Work done during an internship at Salesforce AI Research.} \,\, Jesse Vig$^2$ \,\, Mohit Bansal$^{1}$ \,\, \textbf{Shafiq Joty}$^2$ \\
$^1$UNC Chapel Hill \quad $^2$Salesforce AI Research\\
\texttt{\{davidwan,mbansal\}@cs.unc.edu} \\
\texttt{\{jvig, sjoty\}@salesforce.com}
}

\begin{document}
\maketitle
\begin{abstract}
Large Language Models (LLMs) often exhibit positional bias in long-context settings, under-attending to information in the middle of inputs. We investigate the presence of this bias in long-form summarization, its impact on faithfulness, and various techniques to mitigate this bias. To consistently evaluate faithfulness, we first compile a benchmark of eight human-annotated long-form summarization datasets and perform a meta-evaluation of faithfulness metrics. We show that LLM-based faithfulness metrics, though effective with full-context inputs, remain sensitive to document order, indicating positional bias. Analyzing LLM-generated summaries across six datasets, we find a "U-shaped" trend in faithfulness, where LLMs faithfully summarize the beginning and end of documents but neglect middle content. Perturbing document order similarly reveals models are less faithful when important documents are placed in the middle of the input. We find that this behavior is partly due to shifting focus with context length: as context increases, summaries become less faithful, but beyond a certain length, faithfulness improves as the model focuses on the end. Finally, we experiment with different generation techniques to reduce positional bias and find that prompting techniques direct model attention to specific positions, whereas more sophisticated approaches offer limited improvements. Our data and code are available in \url{https://github.com/meetdavidwan/longformfact}.
\end{abstract}

\begin{figure}
    \centering
    \includegraphics[width=0.7\columnwidth]{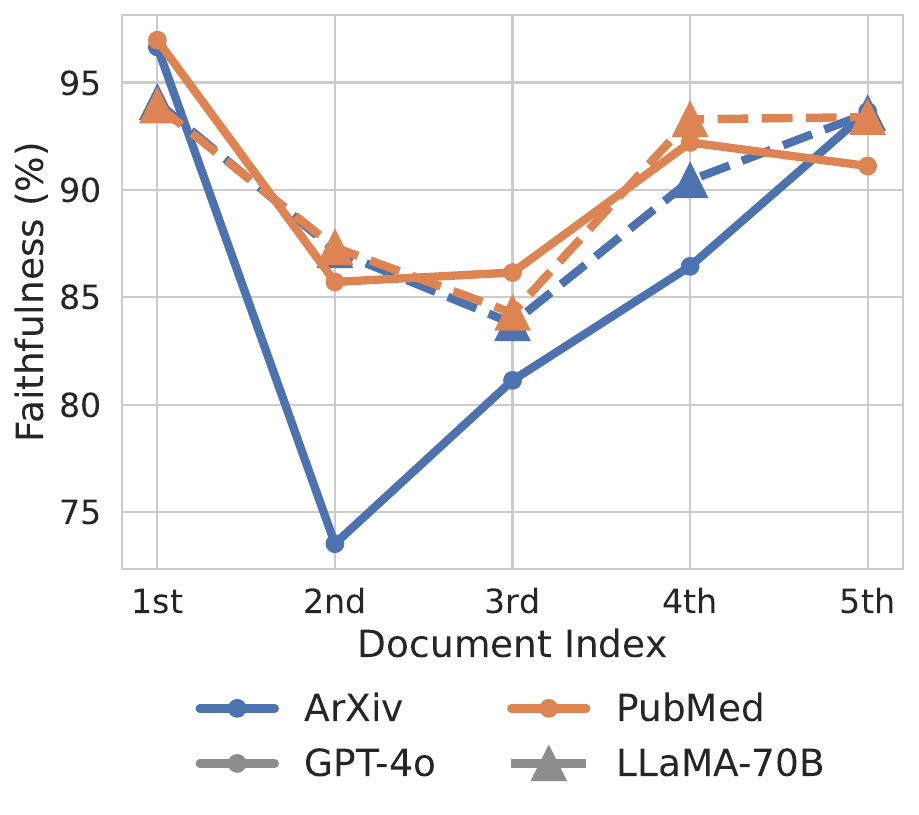}
    \caption{Positional bias in long-form summarization: On two representative models and datasets, summaries are less faithful to the documents in the middle.}
    \label{fig:teaser}
\end{figure}

\begin{table*}[ht!]
    \centering
    \small
    \begin{tabularx}{\textwidth}{X}
    \toprule
    \textbf{RQ1: How to measure faithfulness for long-form summarization?}\\
    \midrule
    - LLM-based metrics excel at using full context.\\
    - Taking the maximum faithfulness score over documents and average over summary sentences is the best merging strategy. \\
    - LLM-based metrics are sensitive towards different orders of documents.\\
    \midrule
    \textbf{RQ2: How faithful are the summaries with respect to the input documents? Are they prone to positional bias?}\\
    \midrule
    - Summaries show lead or lost-in-the-middle bias for faithfulness.\\
    - LLMs are sensitive towards perturbation of documents.\\
    - LLM exhibit a shift of focus with different lengths of the input source, summarizing information towards the end more faithfully when encountering long context.\\
    \midrule
    \textbf{RQ3: How to reduce such positional bias?}\\
    \midrule
    - Simply prompting the models to focus on the different locations of the documents is moderately effective.\\
    - Performing methods that changes the input structure (i.e., hierarchical merging or incremental updating) hurts faithfulness.\\
    \bottomrule
    \end{tabularx}
    \caption{Summary of our research questions and key findings.
    }
    \label{tab:key_findings}
\end{table*}

\section{Introduction}
Large language models (LLMs) have enabled high-quality summary generation. However, the use of LLMs for long-context scenarios, where either the source document(s) or the generated summary is very long, still remains challenging \cite{chang2024booookscore, fables-2024-kim-et-al}. A recent line of work has identified a problem with LLMs, positional bias, where models attend less to relevant information in the middle \cite{liu-etal-2024-lost}. This ``lost-in-the-middle'' trend has been observed beyond long-form question-answering and for summarization \cite{ravaut-etal-2024-context}, where LLMs do not utilize information from the middle of the documents.

One question that arises from such studies is: How does this affect faithfulness? Previous studies have shown that models hallucinate when they are uncertain in their responses \cite{cao-etal-2022-hallucinated, van-der-poel-etal-2022-mutual}. From the ``lost-in-the-middle'' finding, it can be inferred that the weak attention towards the middle context should also lead to hallucinations when generating content about that part. In this paper, we take a step further and analyze the relationship between faithfulness and positional bias for long-form summarization, as summarized in \autoref{tab:key_findings}. We focus on the following research questions:
(1) \textbf{What is the best configuration of LLM-based faithfulness metric for long-form summarization, and how does positional bias affect the metrics?}
(2) \textbf{Are LLM-generated summaries prone to positional bias?}
(3) \textbf{What methods can reduce positional bias?}

For our analysis of faithfulness in long-form document summarization, we first need to determine the best faithfulness metrics, as very few studies have performed an extensive study. Specifically, we want to verify whether current automatic metrics can evaluate summaries well given the large context. And if the metric needs to break the input context into chunks, what is the best way to merge the faithfulness scores. To do so, we collate a large, unified benchmark \benchmark{} for evaluating the performance of metrics on long-form summarization across 8 human-annotated benchmarks. We evaluate LLM-based metrics and find that the models handle full context well, and for the splitting case, taking the maximum over the source documents and taking the average across the summary sentences yield the highest correlation with human judgments. We further perform a perturbation experiment for the full-context setting, where we sort the documents according to their similarity with the summary. An ideal metric should not be affected by the order of the documents, but we find that the metrics are sensitive towards order perturbation and thus suffer from positional bias.

After determining the faithfulness metrics for long-form summarization, we use it to perform an extensive analysis of the faithfulness of the generated summaries across the input documents. Across 6 datasets, we generate summaries and plot the faithfulness scores to verify whether the model hallucinates more for the documents in the middle. Similar to \citet{ravaut-etal-2024-context} that analyzes context utilization for summarization, we observe a U-shape and lead bias when analyzing the faithfulness of the generated summaries. Next, we perform a similar order perturbation experiment and find that LLMs are sensitive to the order, summarizing documents more faithfully for the documents at the beginning. Lastly, we analyze how length correlates with faithfulness by measuring how faithfulness changes as the input length increases. We find that models gradually introduce more hallucinations as we introduce more documents, and after a certain threshold, the model becomes more faithful as it attends to the documents at the end.

Finally, we investigate methods that attempt to mitigate positional bias. We explore methods such as prompting to focus on certain parts, hierarchical merging, incremental updating, and calibration methods. We find that prompting methods are moderately effective at improving summary's faithfulness towards certain positions, while more sophisticated methods struggle to address this issue.

\begin{table*}[!t]
    \centering
    \resizebox{\textwidth}{!}{%
        \begin{tabular}{l| rrrrrcl }
        \toprule
        Tasks & Num Ex. & Doc Split & Doc Words & Summ Split & Summ Words & ann. level & Models \\
        \midrule
        MultiNews & 90 & 3.4 & 767.2 & 7.1 & 175.0 & summ & GPT-3.5, UniSumm, PEGASUS \\
        QMSumm & 90 & - & 1252.8 & 3.04 & 69.2 &  summ & GPT-3.5, UniSumm, PEGASUS \\
        GovReport & 147 & - & 2353.0 & 14.5 & 449.2 & sent & PEGASUS, BART \\
        PubMed & 40 &  6.9 & 3299.2 & 10.4 & 195.0 &  sent & BART, BARTDPR \\
        ArXiv & 146 & 5.6 & 4805.6 & 6.4 & 164.9 & sent & PEGASUS, BART \\
        SQuALITY & 40 & - & 5946.4 & 18.9 & 387.9 & sent & BART, BARTDPR \\
        ChemSumm & 90 &  15.4 & 5974.5 & 7.2& 197.7 &  summ & LongT5, PRIMERA \\
        Diversesumm & 377 & 10.0 & 7644.1 & 7.6 & 203.3 & sent & GPT-4, GPT-3.5, Vicuna, LongChat\\
        \bottomrule
        \end{tabular}
    }
    \caption{Faithfulness meta-evaluation statistics. Ann. level indicates the granularity of the faithfulness annotation.}
    \label{tab:metaeval_statistics}
\end{table*}

\section{Preliminaries}

\subsection{Long-form Summarization}
We consider the task of summarization, where a model generates an $m$-sentence summary $S=\{s_{1}, s_{2}, ..., s_{m}\}$ from input document(s) $D$. For long-form summarization tasks, we follow \citet{ravaut-etal-2024-context} and generally consider that the documents need to contain at least 2k tokens. For this task, such as multi-document summarization, the input $D$ consists of $n$ documents: $D = \{d_{1}, d_{2}, ..., d_{n}\}$. We refer to the boundaries of these documents as \textit{natural document boundaries}, since there are no restrictions on how long each document may be. To unify different datasets, we can similarly split a single-document dataset, such as a scientific document, into different sections and refer to these sections as ``documents.'' Alternatively, one may split the document into fixed-length chunks of words or tokens. To generate summaries, the entire input is truncated to fit the context window of each respective model.

\subsection{Generating Long-form Summaries}
In standard generation, the model $M_{g}$ processes the documents $D$ to produce the output summary $S$, as represented by $M_{g} (D) = S$. To help the model recognize document boundaries, a special indicator is usually inserted between each document; the most commonly used indicator is ``====''. Unless otherwise stated, we use this basic generation setup in most cases. In Section~\ref{sec:reducing_positional_bias}, we further explore other more advanced generation techniques.

\subsection{Faithfulness Evaluation}
For faithfulness evaluation, we consider entailment-based metrics that predict a binary faithfulness label given the document and the generated summaries. In the simplest form, we can make use of the full input $M_{e}(D,S) \in \{0,1\}$. However, due to the prohibitive context length, many metrics that do not have such a large context window require either truncating the input—which loses information crucial for faithfulness evaluation—or splitting the task into evaluations of different document chunks and summaries. Thus, we evaluate both $D$ and $S$ in its more fine-grained form, $M_e(d_{i}, s_{j})$ for the $i$th document chunk and $j$th summary sentence. Finally, to combine the scores, we explore three aggregation methods for both documents and summary sentences: Taking the maximum, minimum, or average $AGG \in \{ max{}, min, mean \}$. Thus, the final score is $\aggs^{n}_{j=0} \aggd^{m}_{i=0} M_{e}(d_{i},s_{j})$.

\section{Faithfulness Metrics Meta-Evaluation}\label{sec:metric}

Given the limited studies in determining the best automatic faithfulness metric for long-form summarization, we first aim to comprehensively test the best strategy for applying current evaluation methods in the long-form context.

\begin{table*}[!th]
    \centering
    \small
    \begin{tabular}{c c | ccccccccc }
    \toprule
    Metric & Doc Merge & MN & QM & GR & PB & AX & SQ & CS & DS & Average \\
    \midrule
     \multirow{4}{*}{MiniCheck} & Original & 53.8 & \underline{53.9} & \underline{50.0} & \underline{79.7} & 50.0 & \textbf{83.0} & 50.9 & \textbf{55.2} & \underline{59.6} \\
    & Min & 54.7 & \textbf{55.7} & \textbf{55.1} & 50.0 & 50.0 & \textbf{83.0} & \underline{54.2} & \underline{49.6} & 56.5 \\
    & Mean & \textbf{64.5} & \textbf{55.7} & \textbf{55.1} & 52.6 & \underline{53.4} & \textbf{83.0} & \underline{54.2} & 46.0 & 58.1 \\
    & Max & \underline{56.6} & \textbf{55.7} & \textbf{55.1} & \textbf{84.8} & \textbf{62.3} & \textbf{83.0} & \textbf{59.7} & 49.3 & \textbf{63.3} \\
    \midrule
     \multirow{4}{*}{GPT-4o}  & Full  & \underline{52.2} & \textbf{64.3} & \textbf{61.6} & \textbf{80.1} & \textbf{57.6} & \underline{76.7} & \textbf{63.4} & \textbf{57.4} & \textbf{64.1} \\
     & Min & 50.0 & \underline{60.7} & \underline{60.0} & 54.4 & 52.1 & \textbf{80.6} & 54.2 & 50.3 & 57.8 \\
     & Mean  & 46.3 & \underline{60.7} & \underline{60.0} & 64.3 & \underline{56.2} & \textbf{80.6} & 54.2 & \underline{56.6} & 59.9 \\
     & Max & \textbf{60.0} & \underline{60.7} & \underline{60.0} & \underline{76.0} & 55.6 & \textbf{80.6} & \underline{60.6} & 53.0 & \underline{63.3} \\
    \bottomrule
    \end{tabular}
    \caption{BACC on meta-evaluation benchmarks, where MN refers to MultiNews, QM to QMSumm, PB to PubMed, GR to GovReport, AX to ArXiv, SQ to SQuALITY, CS to ChemSumm,  and DS to Diversesumm. Min, mean, and max represent the respective operations to merge the document-level faithfulness labels, while full predicts faithfulness with all documents. For MiniCheck, we report the original method by the authors, which internally performs document chunking.
    We \textbf{bold} the best merging strategy for each metric and \underline{underline} the second-best.}
    \label{tab:metaeval_results}
\end{table*}

\subsection{\benchmark{}}\label{sec:benchmark}
To better evaluate faithfulness metrics, we collate a large, unified benchmark consisting of eight long-form summarization datasets. We extend the effort by \citet{zhang-etal-2024-fine} by including additional important long-form summarization annotation \cite{huang-etal-2024-embrace, krishna-etal-2023-longeval}. Statistics about the datasets are reported in \autoref{tab:metaeval_statistics}. Similar to prior unifying efforts \cite{laban-etal-2022-summac,tang-etal-2024-minicheck,zhang-etal-2024-fine}, we convert different annotation schemes into binary faithfulness judgments. For Likert-based evaluations, we consider a summary to be faithful only if it receives the highest score. We describe the datasets below and include more details in Appendix~\ref{sec:dataset_details}:

\paragraph{MultiNews \cite{fabbri-etal-2019-multi}} is a large multi-document news summarization dataset. \citet{chen-etal-2023-unisumm} collected 90 examples with Likert faithfulness scores at summary level.

\paragraph{QMSUM \cite{zhong-etal-2021-qmsum}} is a query-based, multi-domain meeting summarization dataset. \citet{chen-etal-2023-unisumm} similarly collected 90 examples with summary-level Likert faithfulness scores.

\paragraph{ArXiv \cite{cohan-etal-2018-discourse}} is a summarization dataset of scientific articles. \citet{koh-etal-2022-far} collected 146 examples by asking whether each summary sentence contains faithfulness errors. 

\paragraph{GovReport \cite{huang-etal-2021-efficient}} consists of long reports from government research agencies. \citet{koh-etal-2022-far} collected 147 sentence-level annotations for faithfulness.

\paragraph{ChemSumm \cite{adams-etal-2023-desired}} is a scientific long-form summarization dataset in the chemical domain. The authors collected summary-level faithfulness annotations represented by binary labels.

\paragraph{PubMed \cite{cohan-etal-2018-discourse}} is a scientific long-form summarization dataset in the medical domain. \citet{krishna-etal-2023-longeval} collected sentence-level binary judgments of faithfulness. We use the \textsc{fine} annotations, containing faithfulness judgments for each summary sentence. 

\paragraph{SQuALITY \cite{wang-etal-2022-squality}} is a question-focused, long-document summarization dataset, where the documents are short stories from Project Gutenberg. \citet{krishna-etal-2023-longeval} similarly collected sentence-level binary judgments
of faithfulness. We use the \textsc{fine} annotations. 

\paragraph{DiverseSumm \cite{huang-etal-2024-embrace}} is a multi-document news summarization dataset that focuses on conflicting information. The authors collected sentence-level binary faithfulness judgments.

\subsection{Experimental Setup}\label{sec:metric_experimental_setup}

\paragraph{Evaluation Strategy.} Recognizing that models may exhibit positional bias, we experiment with assessing the faithfulness of the summary with respect to each document individually, as well as using the full input source. We also decompose the summary into individual sentences, a technique proven effective by \citet{huang-etal-2024-embrace}. We explore applying minimum, mean, and maximum aggregation methods over the documents. For summary sentence merging, we report mean, and report the result of different merging strategies in Appendix~\ref{sec:metric_summary_aggregation}. We use the \textit{natural document boundaries} to separate the documents, and explore chunking the input into fixed number of tokens in Appendix~\ref{sec:metric_results_fixed_context_window}.

\paragraph{Evaluation Models.} We primarily experiment using GPT-4o as the backbone LLM for the metric. We also explore the applicability of MiniCheck\footnote{We use \textit{Bespoke-MiniCheck-7B}.} \cite{tang2024minicheck} -- an automatic metric that has demonstrated efficacy comparable to GPT-4 performance -- to the long-context setting. We note that MiniCheck by default splits the input into chunks of fixed number of tokens and takes the maximum over the documents, which we include as one of the baselines. Additionally, we present results using the Llama-3.1-8B model in Appendix~\ref{sec:metric_results_full}.

\paragraph{Metric.} To account for class imbalance, we use balanced accuracy (BACC) to calculate metrics' correlations with human judgments.

\subsection{Results}\label{sec:metric_result}
The main results are shown in \autoref{tab:metaeval_results}. For MiniCheck, we observe that taking the maximum over the document achieves the highest correlations on average. This shows that this document aggregation method achieves the best results even for long-form summarization. Interestingly, the original strategy of taking the maximum over fixed input context performs on average $3.7\%$ lower than using natural document boundaries, suggesting that only evaluating the relevant context is important. We also note that MiniCheck trails the strongest GPT-4o-based metric by only $0.8\%$, while matching the performance of GPT-4o-based metric with the same aggregation method.

When looking at the GPT-4o-based metric, using the full context performs the best, achieving the highest accuracy in 5 out of 8 cases and second-best accuracy in 2 of the remaining 3 cases. This suggests that LLMs can utilize long context effectively for evaluation. The second-highest ranking evaluation strategy is still merging the documents by taking the maximum over the documents, performing on average only $0.3\%$ lower than using the full context, aligning with MiniCheck results and previous findings on the best merging strategy.

\begin{table}[!t]
    \centering
    \resizebox{\columnwidth}{!}{
    \begin{tabular}{cc cccc c}
    \toprule
     & & \multicolumn{4}{c}{Document order} \\
     \cmidrule{3-6}
    Dataset & Random? & Original & Top & Middle & Bottom & Sensitivity \\
    \midrule
    ArXiv & \ding{53} & \textbf{57.6} & 56.6 & 55.1 & 57.0 & 2.5 \\
    ChemSumm & \ding{53} & \textbf{63.4} & 61.6 & 57.4 & 59.7 & 6.0\ \\
    PubMed & \ding{53} & 80.1 & 83.2 & 83.9 & \textbf{85.5} & 5.4 \\
    MultiNews &  \ding{51} & 52.2 & \textbf{60.0} & 56.7 & 56.7 & 7.8\\
    DiverseSumm & \ding{51} & \textbf{57.4} & 55.0 & 55.4 & 56.5 & 2.4 \\
    \midrule
    Avg. Sensitivity & - & - & 3.2 & 3.8 & 3.0 & -\\
    \bottomrule
    \end{tabular}
    }
    \caption{BACC using GPT-4o-based metric when order of the documents are perturbed. `Random' indicates whether the initial document orders are random.}
    \label{tab:metaeval_perturbation_results}
\end{table}

\paragraph{Perturbed Document Order.} In addition to standard meta-evaluation, we also perform an analysis by ordering the documents in terms of importance. Specifically, for datasets where document boundaries exist, we calculate the importance of each document relative to the model-generated summary using sentence similarity.\footnote{We use SentenceTransformer \cite{reimers-gurevych-2019-sentence} with the \textit{all-mpnet-base-v2} model.} We then order the documents into top (beginning), middle, and bottom (end), corresponding to the placement of the most important documents. To illustrate, assume there are five documents with importance ranks of 1, 3, 2, 5, and 4, where rank 1 denotes the most important and rank 5 the least. The "top" ordering would sort documents by importance (e.g., 1-2-3-4-5), the "bottom" ordering would prioritize the least important documents (e.g., 5-3-4-2-1), and the "middle" ordering would reflect a mid-tier arrangement (e.g., 4-2-1-3-5). 
Note that the only change from the regular case is the document order, which should have no effect for MultiNews and DiverseSumm, where the documents are in random order, but may have an effect for the other datasets where it breaks the natural flow of the document.The results are presented in \autoref{tab:metaeval_perturbation_results}. In addition to BACC, we also include sensitivity, defined as the maximum difference between scores computed using the original ordering and those with different orderings.

Overall, we find that the metric is sensitive to document order, with high sensitivity observed across each dataset and reaching a 7.8\% difference for MultiNews. When comparing only the top, middle, and bottom orderings, we find no clear trend across datasets; however, on average, the sensitivity is highest when the important document is placed in the middle. This indicates that the metric achieves the lowest BACC when the important document is in the middle, whereas placing it at the top results in the smallest difference, suggesting that the LLM has a stronger lead bias, i.e. performing better when the important documents are at the beginning of the input. Therefore, it may still be beneficial to use the metric that evaluates each document individually and aggregates the results via the maximum operation to reduce positional bias. This approach achieves a similar BACC compared to the full-context setting while inherently not being sensitive to input order.

\paragraph{Takeaway.} We demonstrate that while the GPT-4o based metric is able to utilize the full context, the model exhibits a ``lost-in-the-middle'' behavior when using an LLM as the metric. Therefore, we recommend evaluating each document individually and taking the maximum faithfulness score.

\begin{figure*}[t!]
    \centering
    \includegraphics[width=.95\textwidth]{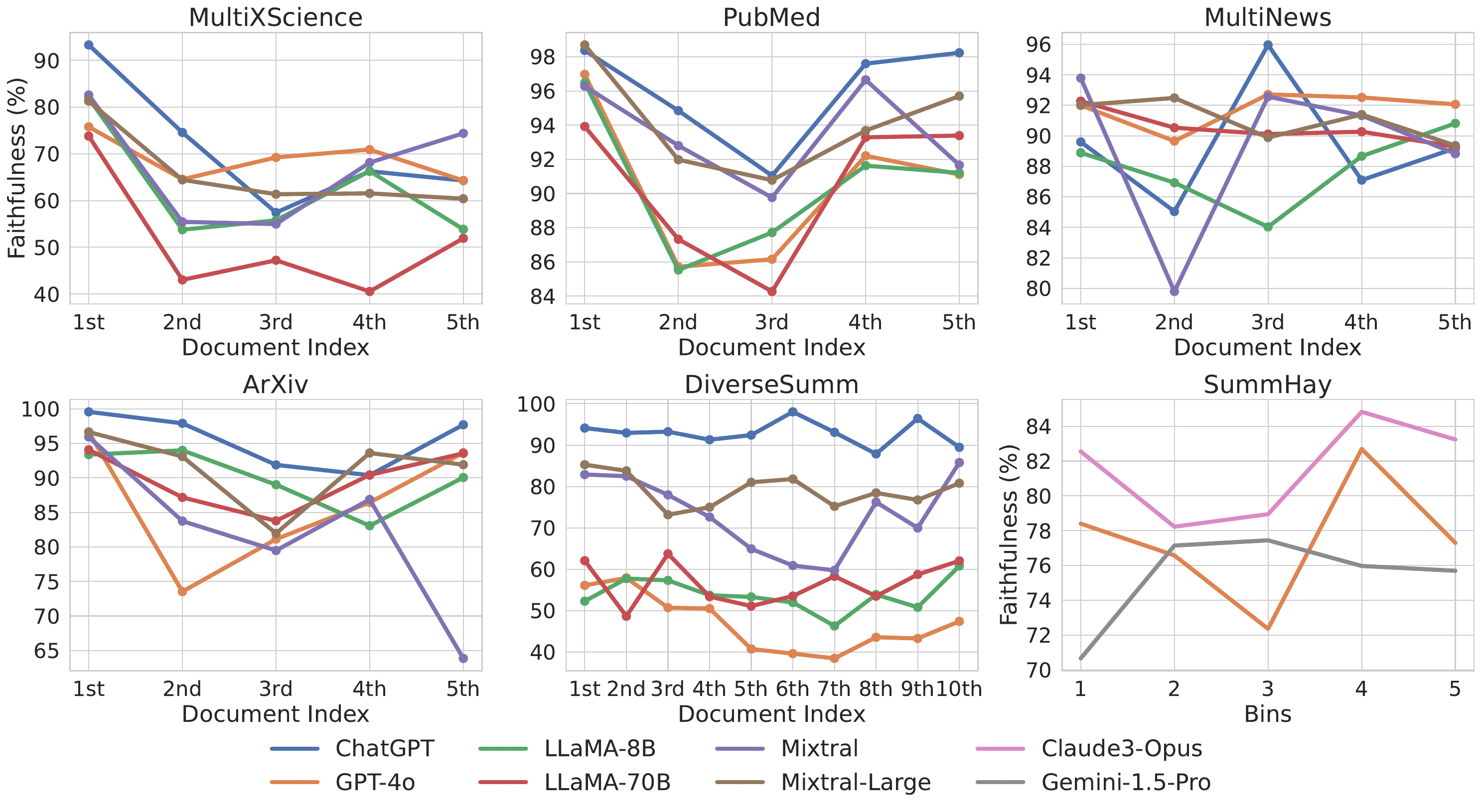}
    \caption{Faithfulness analysis across different positions of documents.}
    \label{fig:direct_analysis}
\end{figure*}

\begin{table}[!t]
    \centering
    \resizebox{\columnwidth}{!}{%
        \begin{tabular}{l| rrrrrcl }
        \toprule
        Tasks & Doc Split & Doc Words & Summ Split & Summ Words\\
        \midrule
        MultiXScience & 5 & 804.9 & 6.9 & 186.0 \\
        PubMed & 5 & 2850.3 & 7.4 & 190.1 \\
        MultiNews & 5 & 4925.5 & 8.5 & 215.3 \\
        ArXiv & 5 & 5825.5 & 7.1 & 181.6 \\
        DiverseSumm & 10 & 7561.5 & 16.0 & 452.5 \\
        SummHay & 100 & 87913.1 & 7.5 & 52.2 \\
        \bottomrule
        \end{tabular}
    }
    \caption{Statistics for the generated summaries.}
    \label{tab:generation_statistics}
\end{table}

\section{Faithfulness of Long-form Summaries}\label{sec:analysis}
Next, we evaluate the faithfulness of summaries generated from different datasets, and perform detailed faithfulness analysis, including assessing faithfulness across each document, performing a perturbed analysis in which we sort documents by importance, and understanding how faithfulness changes as the number of documents increases.

\subsection{Experimental Setup}\label{sec:faithfulness_experimental_setup}
\paragraph{Datasets.} We include two representative multi-document summarization datasets, MultiNews \cite{fabbri-etal-2019-multi} and MultiXScience \cite{lu-etal-2020-multi-xscience}; two long-form summarization datasets, ArXiv and PubMed \cite{cohan-etal-2018-discourse}; and two recent summarization datasets with extremely long contexts, DiverseSumm \cite{huang-etal-2024-embrace} and SummHay \cite{laban2024SummHay}. For ArXiv, PubMed, MultiNews, and MultiXScience, we randomly sample 100 examples from the validation set, each consisting of five documents or sections. For DiverseSumm, we use all original 10 documents and randomly sample 100 examples. The dataset statistics are shown in \autoref{tab:generation_statistics}.

\paragraph{Models.} To comprehensively evaluate positional bias across a range of models, we run GPT-3.5, GPT-4o \cite{gpt4o}, Llama-3.1-8B and 70B \cite{dubey2024llama3herdmodels}, and Mixtral-7×8B and Mixtral-8×22B \cite{jiang2024mixtralexperts}. For SummHay, due to the high computational cost, we reuse the provided generated summaries from GPT-4o, Claude-Opus, and Gemini-1.5-pro.

\paragraph{Evaluation Metrics.} Although GPT-4o demonstrates better faithfulness performance in Section~\ref{sec:metric_result}, we choose MiniCheck for its nearly comparable performance, as well as its greater efficiency and lower cost. In Appendix~\ref{sec:faithfulness_analysis_gpt4o},  we show evaluating with GPT-4o-based metric on a small subset, which exihibits similar trends.

\subsection{Faithfulness Analysis}\label{sec:faithfulness_analysis}

Our main analysis evaluates whether the summary is more faithful to documents in certain positions. To do so, we calculate the Minicheck faithfulness score with respect to all documents individually. We note that if we directly report the summary's faithfulness of each document, it may not accurately reflect faithfulness, as it is also confounded by \emph{coverage}, i.e., how much the summary draws on content from each document. In fact, \citet{huang-etal-2024-embrace} use the faithfulness score per document to measure coverage bias, and we provide an analysis of this coverage in Appendix~\ref{sec:coverage_analysis} as well as an analysis of document content overlap in Appendix~\ref{sec:overlap_analysis}. To remove the effect of coverage, we consider \emph{attribution}; that is, determining which document each sentence of the summary is discussing. For each summary sentence, we take the maximum faithfulness score over all documents. This is, in fact, the same process as one of the best document merging strategies reported in Section~\ref{sec:model_length}.

To illustrate, assume we have as input five documents with no overlap, and a summary consisting of five sentences, each sentence perfectly faithful to one of the documents and unfaithful with respect to the others. If we were to calculate the faithfulness score for each document separately, this summary would appear to be only $20\%$ faithful towards each document (since only one sentence is faithful and the other four are not), as the score is misconstrued by coverage. However, if we only take the maximum faithfulness score over the documents, we remove the coverage effect and show that the summary is indeed faithful towards all documents.

In Appendix~\ref{sec:superpal}, we demonstrate that taking the maximum over the faithfulness scores as attribution also exhibits the same trend as using SuperPAL \cite{ernst-etal-2021-summary}, a document-summary sentence alignment method that achieves the best alignment for long-form summarization \cite{krishna-etal-2023-longeval}.

\begin{figure*}[!t]
    \centering
    \includegraphics[width=\linewidth]{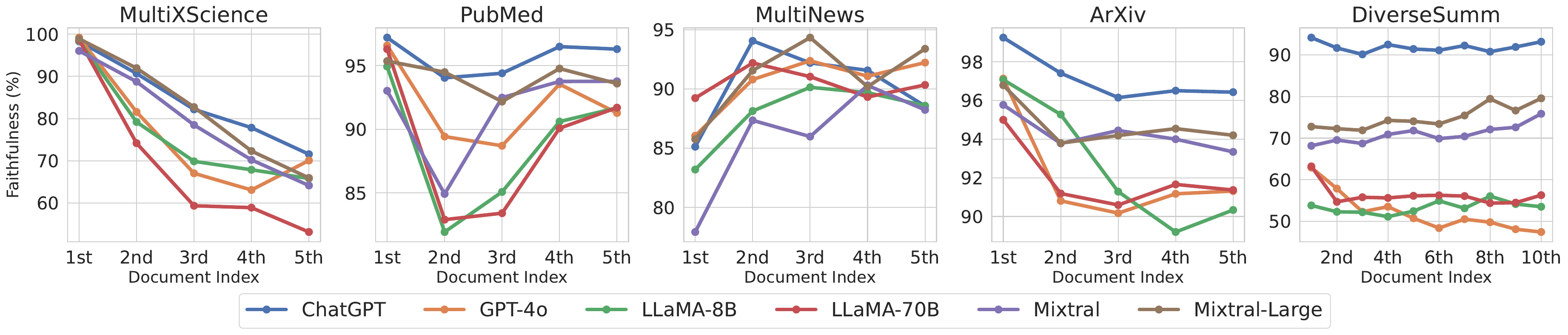}
    \caption{Faithfulness analysis by increasing the context length by adding documents.}
    \label{fig:subset}
\end{figure*}

\paragraph{Results.} We present the results in \autoref{fig:direct_analysis}. Generally, we observe that the models exhibit a dominant U-shaped curve, particularly on ArXiv, PubMed, MultiNews, and MultiXScience, where the middle documents are less faithful than the first or last documents. While most models exhibit this trend, Mixtral shows a more pronounced lead bias, especially on ArXiv. The trend for DiverseSumm is interesting, as it is more of a linear trend. This is partly due to the nature of the task, where we explicitly ask to perform synthesis across the documents, which helps the model to focus on different parts. On SummHay, although the three models behave differently, we observe that GPT-4o and Claude yield lower faithfulness scores for the middle document. Interestingly, Gemini is the only model that exhibits the reverse trend, performing better on the middle documents; however, its average faithfulness score is not as high as that of the other two models. Appendix~\ref{sec:variance} includes further discussions on variance across tasks.

\begin{table}[!t]
    \centering
    \resizebox{\columnwidth}{!}{
        \begin{tabular}{cc cccc c}
        \toprule
        & & \multicolumn{4}{c}{Document order} \\
        \cmidrule{3-6}
        Dataset & random? & Original & Top & Middle & Bottom & Sensitivity\\
        \midrule
        ArXiv & \ding{53} &  91.3 & 91.5 & \textbf{92.2} & 91.7	& 0.9 \\
        PubMed & \ding{53} & 91.3 & 92.6 & \textbf{93.4} & 92.2 & 2.1 \\
        MultiNews & \ding{51}  & \textbf{92.2} & 90.0 & 89.0 & 90.6 & 3.2 \\
        MultiXScience & \ding{51} & \textbf{70.1} & 67.6 & 62.1 & 68.2 & 8.0 \\
        SummHay & \ding{51} & 73.5 & 62.7 & 66.5 & \textbf{84.6} & 11.1 \\
        \midrule
        Avg. Sensitivity & - & - & 3.4 & 4.3 & 3.2 & - \\
        \bottomrule
        \end{tabular}
    }
    \caption{Faithfulness score when perturbing the document order. `Random' indicates whether the initial document orders are random.}
    \label{tab:order_generation}
\end{table}

\subsection{Perturbed Input Analysis}
Next, we conduct a similar perturbation analysis with GPT-4o as described in Section~\ref{sec:metric_result}, where we reorder the documents according to importance. Here, importance is determined using the similarity between each document and the reference summary. We exclude DiverseSumm, as it does not contain reference summaries. The results are reported in \autoref{tab:order_generation}. We observe a similar trend when we analyze the perturbation for the metric: The models generally generate the most faithful summaries either with the original order or when the important document is placed at the front. When looking at the sensitivity across each ordering, the middle case has the highest sensitivity. Interestingly, we observe the bottom ordering achieves a score 21.9 points over the top case for SummHay. We posit that this is because of how models handle long contexts, i.e., focusing towards the end when the context increases, which we confirm in the subsequent analysis on the correlation between increasing context length and faithfulness.

\subsection{Faithfulness and Length Correlation}\label{sec:model_length}
So far, our analyses have been post-hoc, where we attempt to analyze the faithfulness scores when the input is fixed. Here, we try to analyze how faithfulness correlates with length by incrementally increasing the number of documents. For all datasets, we start with summarizing one document, and then we add one more document and summarize again. We then calculate the faithfulness scores of the generated summaries using the corresponding set of documents. We exclude SummHay here, as it is computationally expensive to run the subsets for all 100 documents incrementally.

Results are in \autoref{fig:subset}. In MultiXScience, which contains relatively few words per document, we observe a strong lead bias. However, when moving to long-form summarization on datasets such as Pubmed and ArXiv, we observe a U-shaped trend: summary faithfulness decreases as the number of documents grows, then begins to improve beyond a certain threshold. This suggests that the model may “switch modes” at a particular context length and focus more on documents introduced later. On MultiNews, a similar pattern appears, as faithfulness steadily increases for most models with additional input documents. These observations are consistent with prior research on LLMs’ behavior with extremely long inputs, where models concentrate much of their attention on the later sections \cite{kim2024fables, laban2024SummHay}.

\begin{figure*}[!t]
    \centering
    \includegraphics[width=\linewidth]{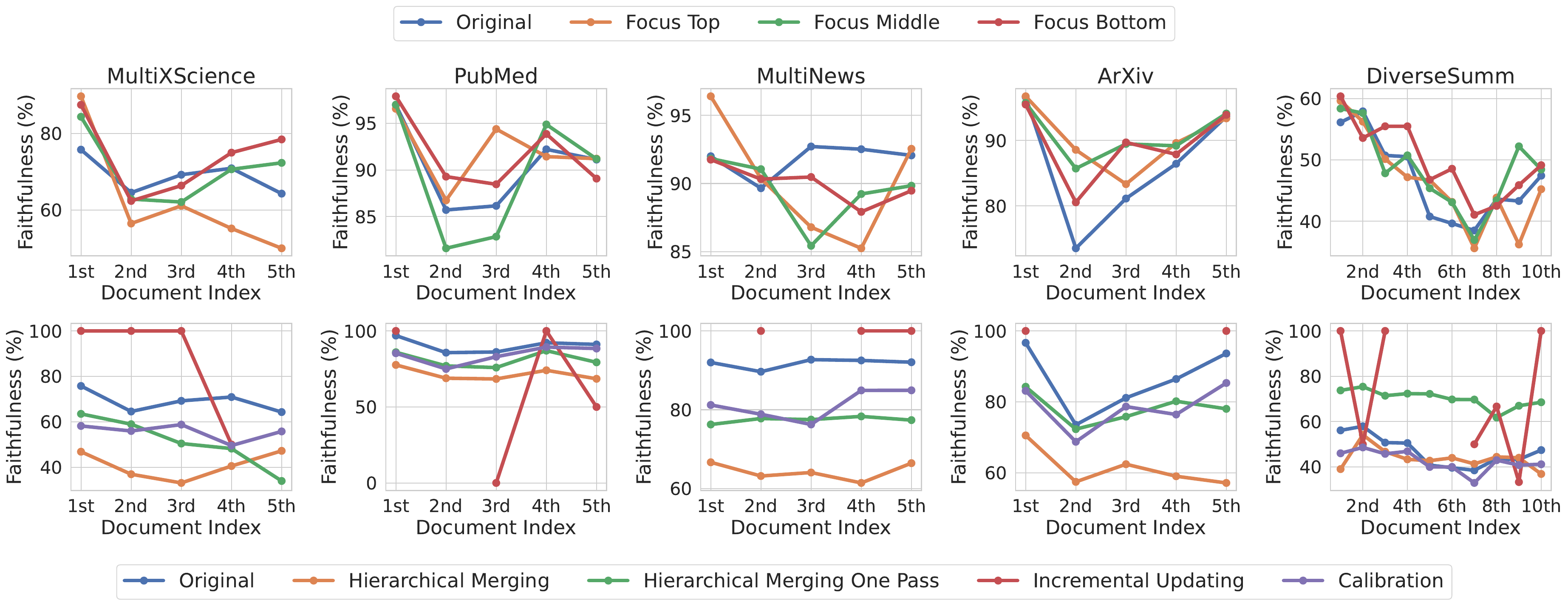}
    \caption{Faithfulness analysis for different summarization techniques.}
    \label{fig:method_faithfulness}
\end{figure*}
\section{Methods for Reducing Positional Bias}\label{sec:reducing_positional_bias}
Finally, we explore different generation methods and verify whether they can reduce positional bias. We use the same sets of examples used in Section~\ref{sec:analysis}.

\subsection{Methods}

\paragraph{Focus Prompt.}
We ask the model to focus on different parts of the input. We append the following instruction to the original prompt: "Focus on the top documents", "focus on the bottom documents", and "focus on the documents in the middle".

\paragraph{Hierarchical Merging.} Explored in \citet{chang2024booookscore}, this method generates one summary for each document, and then iteratively merge the summaries. We explore both merging two summaries at a one until one final summary is produced, as well as one pass over all the individual summaries.

\paragraph{Incremental Updating.} Simiraly explored in \citet{chang2024booookscore}, starting with summarizing the first document, the model updates its output given the current summary and the next document.

\paragraph{Calibration.} Past studies have found that calibrating the model specifically for the positional bias can reduce such bias. For both open-source and propietary models, we generate summary from all permutations of the input order, and then ask the model to combine the generated summaries.\footnote{We also explore logit-level calibration similar to \citet{tang-etal-2024-found}, but this exceeds memory constraints.}

\subsection{Results}
We present the results of running the different methods with GPT-4o in \autoref{fig:method_faithfulness}. The top portion of the figure shows the prompting-based methods. Instructing the model to focus on either the top or bottom documents proves effective for improving faithfulness at those specific positions. For instance, the ``focus top'' prompt yields higher faithfulness scores for the last document than the original prompt in 4 out of 5 datasets. The ``focus middle'' prompt achieves higher faithfulness than the original case for the middle document on ArXiv and DiverseSumm. However, we observe that the model often fails to follow the instructions adequately to focus on the middle documents: ``focus top'' achieves the best faithfulness for the middle documents on PubMed, while ``focus bottom'' performs best on DiverseSumm. This suggests that prompting can partially alleviate positional bias.

Interestingly, the more sophisticated methods illustrated at the bottom of \autoref{fig:method_faithfulness} perform worse than the original prompt across all datasets. It is noteworthy, as we generally observe a very different trend for methods that change the summarization protocol. For example, on ArXiv, PubMed, and DiverseSumm, incremental updating contains gaps in the corresponding lines on the plot. This issue arises because the method maintains only a working summary cache and updates it solely based on the next document, leading to a pronounced recency bias: the model either retains its current summary and thus remains faithful to the first document or focuses on the last document. While this approach may appear to improve faithfulness, it ultimately replaces one positional bias, i.e., lost-in-the-middle, with another, i.e., recency bias.

\section{Related Work}
\paragraph{Summarization.}
As the capabilities of LLMs steadily improve, they exhibit strong performance on traditional summarization tasks \cite{10.1162/tacl_a_00632}, such as XSum \cite{xsum-emnlp} and CNN/Daily Mail \cite{see-etal-2017-get}. Consequently, recent studies have focused on harder tasks to better understand the limitations of LLMs, such as instruction-controllable summarization \cite{liu-etal-2024-benchmarking}, query-focused summarization \cite{tang-etal-2024-tofueval}, summarization of diverse information \cite{huang-etal-2024-embrace,zhang-etal-2024-fair}, and, more recently, long-form summarization \cite{laban2024SummHay, fables-2024-kim-et-al}. Our work focuses on long-form summarization because generating and evaluating summaries with long context lengths is particularly challenging. Therefore, to our knowledge, we present the first study to analyze the faithfulness of generated summaries across a wide range of long-form summarization datasets, employing various generation techniques, and to examine the effect of positional bias on them.

\paragraph{Faithfulness Evaluation Metrics.} To improve the benchmarking of summarization systems, many studies collect human annotations of faithfulness judgments to create meta-evaluation benchmarks that measure the effectiveness of faithfulness metrics \cite{fabbri-etal-2021-summeval, pu2023summarizationalmostdead,tang-etal-2023-understanding,goyal2023newssummarizationevaluationera,10.1162/tacl_a_00632,liu-etal-2023-revisiting,liu-etal-2024-benchmarking}. This effort has led to the development of strong faithfulness metrics \cite{laban-etal-2022-summac, fabbri-etal-2022-qafacteval, zha-etal-2023-alignscore}. More recently, LLM-based metrics, which prompt powerful LLMs to perform evaluations \cite{liu-etal-2023-g, wang-etal-2023-chatgpt, fu-etal-2024-gptscore}, have shown high correlations with human judgments. Subsequently, MiniCheck \cite{tang2024minicheck} has focused on distilling such knowledge into smaller NLI models, combining both strong performance and efficiency. While most of these studies concentrate on standard summarization tasks, few address long-form summarization \cite{krishna-etal-2023-longeval, zhang2024finegrainednaturallanguageinference,huang-etal-2024-embrace}. Specifically, \citet{zhang2024finegrainednaturallanguageinference} collate a meta-evaluation benchmark and develop an automatic metric. We further extend their work by incorporating additional benchmarks, analyzing the performance of the latest faithfulness metrics, and, more importantly, investigating sensitivity to positional bias.

\paragraph{Positional Bias.}
Many works have found that current LLMs exhibit different positional biases. For example, \citet{sun-etal-2021-long} find that models exhibit a recency bias, where the most recent tokens play a stronger role, and that the order of in-context examples significantly affects performance \cite{liu-etal-2022-makes, lu-etal-2022-fantastically, li2024longcontextllmsstrugglelong}. Similarly, such biases also affect LLM performance in arithmetic tasks \cite{shen2023positionaldescriptionmatterstransformers}, multiple-choice questions \cite{llm-mcq-bias, pezeshkpour2023largelanguagemodelssensitivity}, ranking \cite{alzahrani-etal-2024-benchmarks, tang-etal-2024-found}, and evaluation \cite{wang-etal-2024-large-language-models-fair}. Specifically in summarization, \citet{huang-etal-2024-embrace} find that evaluators of faithfulness and coverage highly prefer one choice over another in pairwise settings, and that generated summaries tend to focus on the first and last sections of documents. Additionally, \citet{laban2024SummHay} find that different LLMs exhibit different positional preferences. To analyze this effect more rigorously, \citet{ravaut-etal-2024-context} systematically analyze how positional bias affects context utilization. Our work instead focuses on faithfulness, another crucial aspect of summarization. As discussed in Section~\ref{sec:faithfulness_analysis}, faithfulness is harder to evaluate as it is confounded by coverage and thus requires attributions.

\section{Conclusion}
In this work, we present an extensive analysis of the relationship between positional bias and faithfulness for long-context summarization from three perspectives. We first evaluate the best strategy for assessing faithfulness in long-context summarization tasks, as well as the metrics' sensitivity to positional changes. We find that, although current LLM-based metrics achieve the highest correlation when using the full context, they are sensitive to changes in the order of the input documents.

We then analyze faithfulness of model generations with the best faithfulness metrics. We generate summaries using both open-source and proprietary models and find that the faithfulness of the middle documents tends to dip compared to those at the beginning and end, and the summaries also exhibit high sensitivity when the order of inputs is perturbed. One of the possible explanations, as we find, is that the models change behavior after a certain context length and focus on documents toward the end, improving faithfulness for the documents at the back after the initial dip.

Finally, we investigate several generation methods to test whether they can alleviate positional bias. We find that prompting methods can partially alleviate the middle curse, while more extensive methods provide overall less faithful summaries.

\section*{Limitations}
This work extensively studies the relationship between position and faithfulness in long-context summarization. We acknowledge that there are additional LLMs, such as Claude or CommandR+, and more datasets that could be included in our evaluation. However, due to practical limitations, we have chosen to evaluate a representative and diverse set of LLMs and datasets. Although different models may exhibit varying trends, our analysis reveals that all models exhibit a similar trend regarding positional bias.

Furthermore, although our analysis relies on automatic metrics -- and despite our extensive efforts in Section~\ref{sec:benchmark} to identify the most effective ones -- it may not accurately reflect the trends that would emerge if human annotators evaluated all generated summaries. We do note, however, that human annotations for long-form summarization are both expensive and unreliable due to the extensive context involved \cite{krishna-etal-2023-longeval}. Nevertheless, we hope that our work provides some initial insights into this problem.

In our experiments, we limit the number of documents to five for all datasets (except DiverseSumm) to control for the effect of varying context lengths. Exploring settings with different numbers of documents would be an interesting direction for future work. Nevertheless, we hope that our analysis of faithfulness with different input context lengths sheds light on what we would expect to observe with varying input lengths.

Lastly, we did not rigorously tune all the prompts in Section~\ref{sec:reducing_positional_bias}, which may lead to further improvements in mitigating the middle curse.

We do not forsee any particular risks beyond those inherent to any text generation task. In fact, our work actually focuses on understanding and improving faithfulness for long-form summarization.

\section*{Acknowledgement}
We thank the anonymous reviewers for their valuable feedback. We also would like to thank Philippe Laban and Kung-Hsiang Huang for the helpful discussions and for providing help with the datasets SummHay and DiverseSumm, respectively.

\bibliography{custom}

\appendix

\section{Additional Experimental Setup Details}

\subsection{Dataset Details}\label{sec:dataset_details}
The licenses for the datasets are as follows. QMSUM and MultiNews are under
MIT License. Diversesumm, SummHay, ArXiv, and Pubmed are released under the Appache 2.0 license. SQuality is under the CC BY 4.0 license. GovReport and ChemSumm do not specify any license. We use the authors' original repository and instructions to prepare and process the dataset. The authors of the respective datasets have filtered any harmful content. For annotations, we similarly follow author's instructions to download and process the data.

\subsection{Model Details}
For all models, we use the default generation methods. For open-source models, we use the available Huggingface repository for Llama-3.1-8B\footnote{\url{https://huggingface.co/meta-llama/Llama-3.1-8B-Instruct}} and 70B\footnote{\url{https://huggingface.co/meta-llama/Llama-3.1-70B-Instruct}}, and also for Mixtral\footnote{\url{https://huggingface.co/mistralai/Mixtral-8x7B-Instruct-v0.1}} and Mixtral Large.\footnote{\url{https://huggingface.co/mistralai/Mixtral-8x22B-Instruct-v0.1}} For GPT-4o, we use \textit{gpt-4o} as of October 13th, 2024. We run with \textit{bfloat16} and use 8 A100s to run all generations and evaluations. The approximate costs for GPT-4o are as follows: metric full setting costs \$37.1, metric splitting the document setting costs \$79.9, generation costs \$ 15.2, and generation with different methods costs \$97.2.

\section{Additional Metric Results}\label{sec:metric_summary_aggregation}
\begin{table}[!t]
    \centering
    \resizebox{\columnwidth}{!}{
        \begin{tabular}{c cccccc c}
         \toprule
         Method & MN & QM & GR & AX & CS & DS & Avg. \\
         \midrule
         Full & 52.2 & 64.3 & 61.6 & 57.6 & 63.4 & 57.4 & \underline{59.4} \\
         Natural & 51.9 & 62.5 & 51.5 & 50.0 & 50.0 & 50.6 & 52.8 \\
         \midrule
         Chunk 1024 & 58.1 & 57.1 & 68.5 & 54.7 & 64.8 & 54.3 & \textbf{59.6} \\
         Chunk 2048  & 50.0 & 50.3 & 47.0 & 53.2 & 66.7 & 55.1 & 53.7 \\
         Chunk 4096 & 50.0 & 50.3 & 47.3 & 43.7 & 53.7 & 52.8 & 49.6\\
         Chunk 8192 & 50.0 & 50.0 & 50.0 & 50.7 & 53.2 & 52.3 & 51.0 \\
        \bottomrule
        \end{tabular}
    }
    \caption{BACC of different chunking methods with GPT-4o. We use the best strategy of taking the maximum over documents and average over the summary sentences. As baselines, we report the full input setting and running the same metric with natural document boundaries.}
    \label{tab:metaeval_rensults_chunk}
\end{table}

\begin{table*}[!t]
    \centering
    \resizebox{\linewidth}{!}{%
        \begin{tabular}{c cc | ccccccccc }
        \toprule
        Metric & Doc Merge & Summ Merge & MN & QM & GR & PB & AX & SQ & CS & DS & Avg. \\
        \midrule
        \multirow{12}{*}{MiniCheck} & Original & Min & 47.8 & 56.6 & 64.7 & 79.7 & 76.0 & 83.0 & 46.8 & 53.5 & \textbf{63.5} \\
        & Original & Mean & 56.1 & 55.7 & 55.1 & 79.7 & 57.0 & 83.0 & 57.9 & 54.1 & 62.3 \\
        & Original & Max & 53.8 & 53.9 & 50.0 & 79.7 & 50.0 & 83.0 & 50.9 & \textbf{55.2} & 59.6 \\
        \cmidrule{2-12}
        & Min & Min & 49.4 & 56.6 & 64.7 & 50.0 & 50.0 & 83.0 & 51.4 & 49.8 & 56.9 \\
        & Min & Mean & 54.7 & 55.7 & 55.1 & 50.0 & 50.0 & 83.0 & 54.2 & 49.6 & 56.5 \\
        & Min & Max & 60.3 & 53.9 & 50.0 & 50.0 & 50.8 & 83.0 & 54.2 & 53.8 & 57.0 \\
        \cmidrule{2-12}
        & Mean & Min & 49.4 & 56.6 & 64.7 & 52.6 & 50.0 & 83.0 & 51.4 & 49.3 & 57.1 \\
        & Mean & Mean & 64.5 & 55.7 & 55.1 & 52.6 & 53.4 & 83.0 & 54.2 & 46.0 & 58.1 \\
        & Mean & Max & 60.4 & 53.9 & 50.0 & 52.6 & 46.2 & 83.0 & 66.7 & 51.2 & 58.0\\
        \cmidrule{2-12}
        & Max & Min & 39.4 & 56.6 & 64.7 & 84.8 & 69.7 & 83.0 & 53.2 & 47.1 & 62.3 \\
        & Max & Mean & 56.6 & 55.7 & 55.1 & 84.8 & 62.3 & 83.0 & 59.7 & 49.3 & \underline{63.3} \\
        & Max & Max & 53.2 & 53.9 & 50.0 & 84.8 & 50.0 & 83.0 & 50.9 & \underline{54.2} & 60.0 \\
         \midrule
         \multirow{12}{*}{GPT-4o} & Full & Min & 38.1 & 60.5 & \textbf{64.7} & \textbf{80.1} & \textbf{73.1} & \underline{76.7} & 51.4 & \textbf{60.6} & 63.1\\
         & Full & Mean & 52.2 & 64.3 & \underline{61.6} & \textbf{80.1} & 57.6 & \underline{76.7} & \underline{63.4} & \underline{57.4} & \textbf{64.1} \\
         & Full & Max & 52.2 & 64.3 & \underline{61.6} & \textbf{80.1} & 57.6 & \underline{76.7} & \underline{63.4} & \underline{57.4} & \textbf{64.1} \\
         \cmidrule{2-12}
         & Min & Min & 50.0 & \textbf{65.4} & 58.8 & 54.4 & 50.4 & \textbf{80.6} & 50.0 & 50.2 & 57.5\\
         & Min & Mean & 50.0 & \underline{60.7} & 60.0 & 54.4 & 52.1 & \textbf{80.6} & 54.2 & 50.3 & 57.8\\
         & Min & Max & 50.0 & \underline{60.7} & 60.0 & 54.4 & 52.1 & \textbf{80.6} & 54.2 & 50.3 & 57.8\\
         \cmidrule{2-12}
         & Mean & Min & 46.8 & \textbf{65.4} & 58.8 & 64.3 & 51.7 & \textbf{80.6} & 50.0 & 52.8 & 58.8\\
         & Mean & Mean & 46.3 & \underline{60.7} & 60.0 & 64.3 & 56.2 & \textbf{80.6} & 54.2 & 56.6 & 59.9\\
         & Mean & Max & \underline{55.2} & 59.2 & 50.0 & 64.3 & 49.8 & \textbf{80.6} & \textbf{70.8} & 54.4 & 60.5\\
         \cmidrule{2-12}
         & Max & Min & 44.6 & \textbf{65.4} & 58.8 & 76.0 & \underline{68.8} & \textbf{80.6} & 47.7 & 58.2 & 62.5\\
         & Max & Mean & \textbf{60.0} & \underline{60.7} & 60.0 & 76.0 & 55.6 & \textbf{80.6} & 60.6 & 53.0 & \underline{63.3}\\
         & Max & Max & 51.3 & 59.2 & 50.0 & 76.0 & 50.0 & \textbf{80.6} & 50.0 & 51.5 & 58.6\\
        \midrule
        \multirow{12}{*}{Llama-3.1-8B} & Full & Min & 52.0 & 56.8 & \textbf{64.9} & \textbf{69.9} & 56.3 & \textbf{73.4} & \textbf{67.6} & \underline{52.6} & \textbf{61.7} \\
         & Full & Mean & 53.8 & 55.0 & 51.2 & \textbf{69.9} & 50.0 & \textbf{73.4} & 50.9 & 52.1 & 57.0\\
         & Full & Max & 53.8 & 55.0 & 51.2 & \textbf{69.9} & 50.0 & \textbf{73.4} & 50.9 & 52.1 & 57.0\\
         \cmidrule{2-12}
         & Min & Min & 49.4 & \textbf{63.9} & \underline{63.4} & \underline{63.7} & \textbf{59.5} & \underline{69.9} & 53.2 & 49.1 & 59.0\\
         & Min & Mean & 50.1 & \underline{62.5} & 51.5 & \underline{63.7} & \underline{59.0} & \underline{69.9} & 55.6 & 50.8 & 57.9 \\
         & Min & Max & 50.1 & \underline{62.5} & 51.5 & \underline{63.7} & \underline{59.0} & \underline{69.9} & 55.6 & 50.8 & 57.9\\
         \cmidrule{2-12}
         & Mean & Min & \underline{56.7} & \textbf{63.9} & \underline{63.4} & 63.5 & 51.3 & \underline{69.9} & \underline{61.6} & 51.4 & \underline{60.2} \\
         & Mean & Mean & \textbf{59.3} & \underline{62.5} & 51.5 & 63.5 & 50.0 & \underline{69.9} & \underline{61.6} & 50.9 & 58.7 \\
         & Mean & Max & 49.3 & 52.6 & 50.0 & 63.5 & 50.0 & \underline{69.9} & 50.0 & 49.7 & 54.4\\
         \cmidrule{2-12}
         & Max & Min & 50.9 & \textbf{63.9} & \underline{63.4} & 53.7 & 51.4 & \underline{69.9} & 48.6 & \textbf{53.3} & 56.9\\
         & Max & Mean & 51.9 & \underline{62.5} & 51.5 & 53.7 & 50.0 & \underline{69.9} & 50.0 & 50.6 & 55.0\\
         & Max & Max & 50.0 & 52.6 & 50.0 & 53.7 & 50.0 & \underline{69.9} & 50.0 & 50.0 & 53.3\\
         \bottomrule
        \end{tabular}
    }
    \caption{Full BACC on meta-evaluation benchmarks. CS=ChemSumm, AX=ArXiv, GR=GovReport, QM=QMSumm, MN=MultiNews, DS=Diversesumm. For each metric, best and second-best are bolded and underlined, respectively.BACC on meta-evaluation benchmarks, where MN refers to MultiNews, QM to QMSumm, PB to PubMed, GR to GovReport, AX to ArXiv, SQ to SQuALITY, CS to ChemSumm,  and DS to Diversesumm. We \textbf{bold} the best merging strategy for each metric and \underline{underline} the second-best.}
    \label{tab:metaeval_results_full}
\end{table*}

\subsection{Results on Chunking}\label{sec:metric_results_fixed_context_window}
Instead of using natural document boundaries, we explore chunking the full input into fixed numbers of tokens. Using the same setup as in the meta-evaluation but excluding PubMed and SQuality, we split the input documents into chunks of 1,024, 2,048, 4,096, and 8,192 tokens. We employ the GPT-4-based metric and the best merging strategy (i.e., maximum over documents and average over the summary sentences). The result is shown in \autoref{tab:metaeval_rensults_chunk}. We observe that chunking into smaller chunks (1,024 tokens) leads to stronger performance than using the full context, while other chunk sizes result in worse correlations compared to the full context. This may indicate that smaller context windows help the models, since LLMs still prefer shorter contexts. The fact that natural boundaries are only better than chunking with 4,096 tokens suggests that limiting based on size may be preferable to keeping the original documents as they are, since there is no control over how long each document is.

\subsection{Full Metric Results}\label{sec:metric_results_full}
We present the full results, including an exploration of summary sentence-level merging strategies, in \autoref{tab:metaeval_results_full}. For MiniCheck, the original chunking method using the minimum function over summary sentences yields the highest correlation—though this is largely driven by high accuracy on ArXiv. Meanwhile, the maximum strategy with mean aggregation, which previous studies have identified as most effective, falls short by only 0.2 points. For GPT-4o, we observe that, across all document merging strategies, averaging over the summary sentences achieves the highest correlations overall. This finding is consistent with prior work.

We also run the meta-evaluation with Llama-3.1-8B model in the bottom of \autoref{tab:metaeval_results_full}. Similar to running with GPT-4o, we observe that using the full context achieves the highest correlation on average. Nevertheless, the best summary sentence merging strategy is taking the minimum, and the second-best document merging strategy is mean.

\begin{figure*}[t!]
    \centering
    \includegraphics[width=0.9\textwidth]{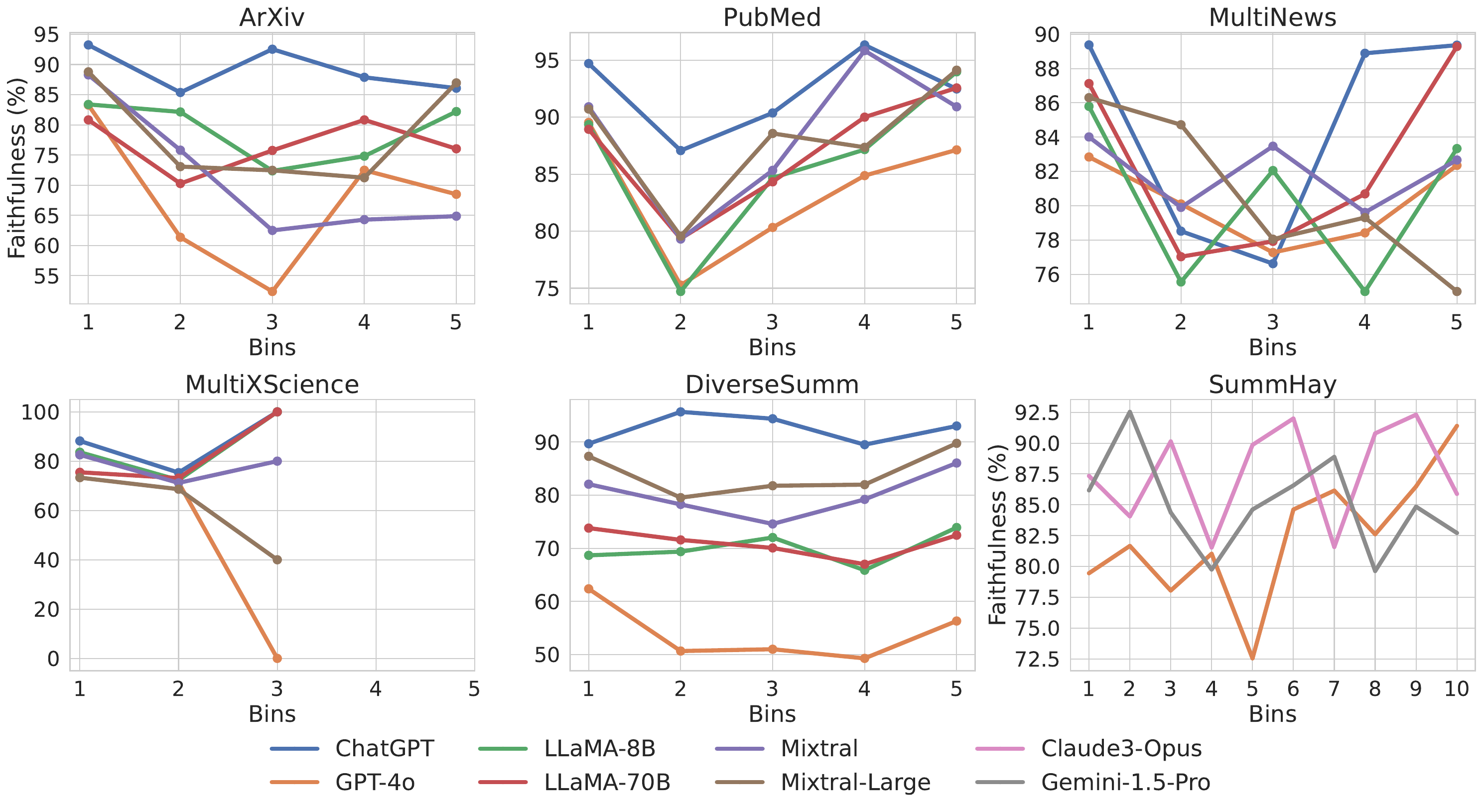}
    \caption{Faithfulness analysis across different positions of documents with chunking.}
    \label{fig:direct_analysis_chunk}
\end{figure*}

\section{Additional Faithfulness Analysis Results}

\begin{table}[!ht]
    \centering
    \begin{tabular}{l ccc}
    \toprule
    Task & R1 & R2 & RL \\
    \midrule
    MultiXScience & 24 & 24 & 12 \\
    PubMed & 26 & 26 & 13 \\
    MultiNews & 18 & 28 & 13 \\
    ArXiv & 32 & 32 & 13 \\
    DiverseSumm & 41 & 41 & 21 \\
    \bottomrule
    \end{tabular}
    \caption{Similarity of the documents using ROUGE.}
    \label{tab:similarity_analysis}
\end{table}
\subsection{Content Overlap Analysis}\label{sec:overlap_analysis}
\begin{figure*}[t!]
    \centering
    \includegraphics[width=\textwidth]{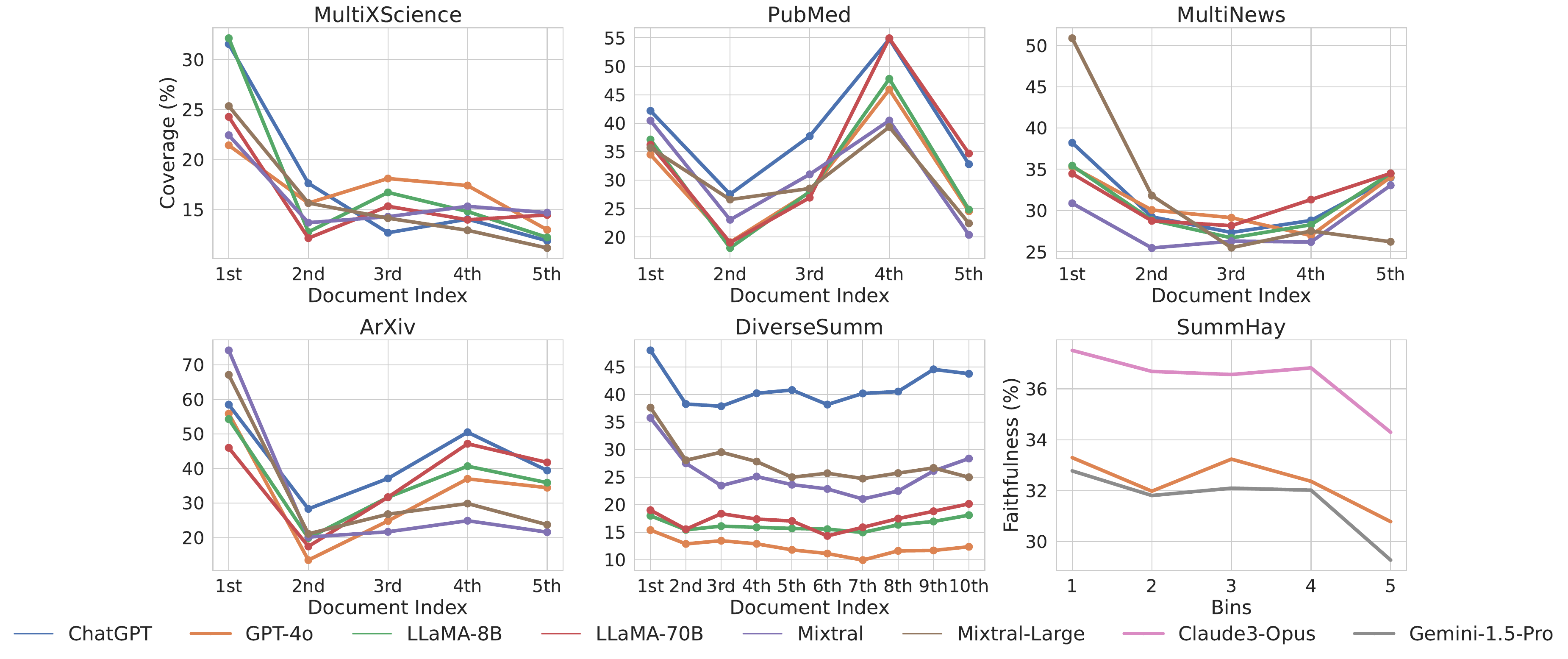}
    \caption{Coverage analysis across different positions.}
    \label{fig:coverage_analysis}
\end{figure*}

Taking the maximum faithfulness implicitly assumes that the conent do not overlap. To verify the variability in similarity, we calculate ROUGE-1/2/L between all document pairs and aggregate the results to show the variablity for each task . We exclude SummHay as it is prohibitively expensive to calculate this for 100 documents. \autoref{tab:similarity_analysis} shows the results. Examining unigrams (R1), we observe a high degree of overlap, particularly for DiverseSumm. This is expected, as all 10 documents focus on the same news event. We note that while such overlap may not be covered by only considering the highest faithfulness for one of the documents, we evaluate faithfulness relative to a single document to minimize the influence of coverage, as mentioned in Section~\ref{sec:faithfulness_analysis}. Determining how many documents a summary can be attributed to is complex and would require defining a threshold, which we leave for future work. Nevertheless, the coverage analysis shown in \autoref{fig:coverage_analysis} can be considered the case where the summary is related to all documents.

\subsection{Faithfulness Analysis with Chunking}
We perform the analysis with calculating the faithfulness with chunking, instead of natural boundaries. We show it in \autoref{fig:direct_analysis_chunk}.

\begin{figure*}[t!]
    \centering
    \includegraphics[width=\textwidth]{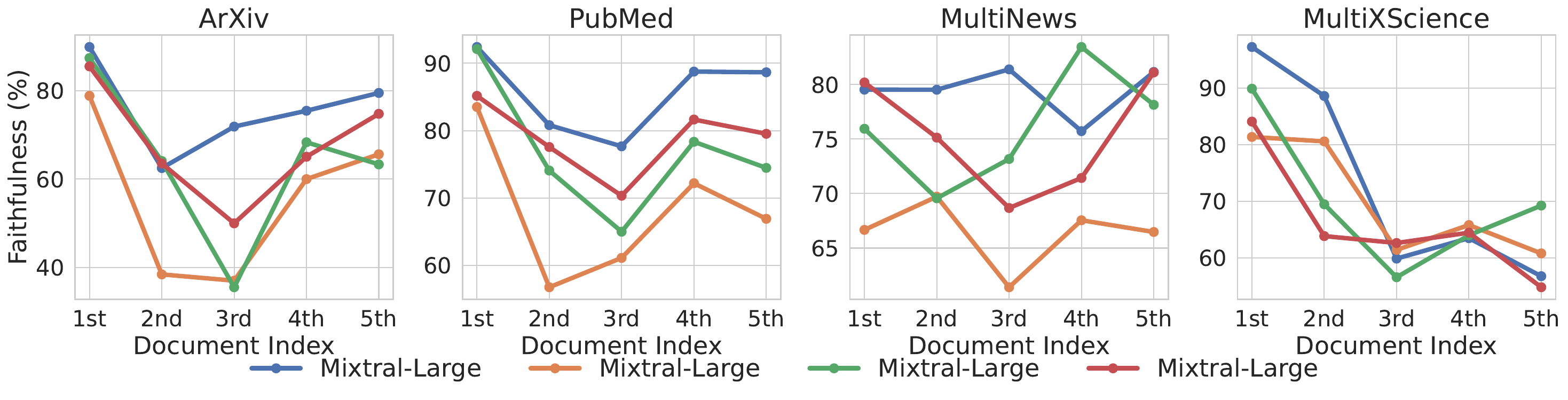}
    \caption{Faithfulness analysis across different positions with SuperPal as alignment.}
    \label{fig:direct_analysis_superpal}
\end{figure*}

\subsection{Discussions on Variance Across Tasks}\label{sec:variance}
In \autoref{fig:direct_analysis}, we observe a large variance across the different tasks. We provide a discussion on possible reasons.

\paragraph{Context Length.} Based on the dataset statistics provided earlier, we analyzed the tasks from those with the shortest document lengths to those with the longest. This reveals an intriguing trend: MultiXScience, shown in \autoref{fig:direct_analysis}, exhibits primarily a lead bias due to its short input length. As we move to datasets with longer input lengths, the U-shaped trend becomes more apparent. MultiXScience can thus be interpreted as representing the early stage of this trend, which evolves as context length increases. We demonstrate this similar progression in Section~\ref{sec:model_length}, where increasing contexts shift the observed behavior. The downward trend of MultiXScience in \autoref{fig:subset} can similarly be attributed to insufficient context length, leaving it in the early lead bias phase.

\paragraph{Task Type.} Another observable trend is the variation across different types of summarization tasks. For example, ArXiv and PubMed involve single long-document summarization, whereas MultiNews, MultiXScience, DiverseSumm, and SummHay are classified as MDS. Our findings indicate that single-document summarization tasks exhibit more consistent model behavior and lower variance, while MDS tasks show much higher variance. We hypothesize that this increased variability stems from the need for synthesis across multiple documents, which impacts the faithfulness of generated summaries.

\begin{figure*}[t!]
    \centering
    \includegraphics[width=\textwidth]{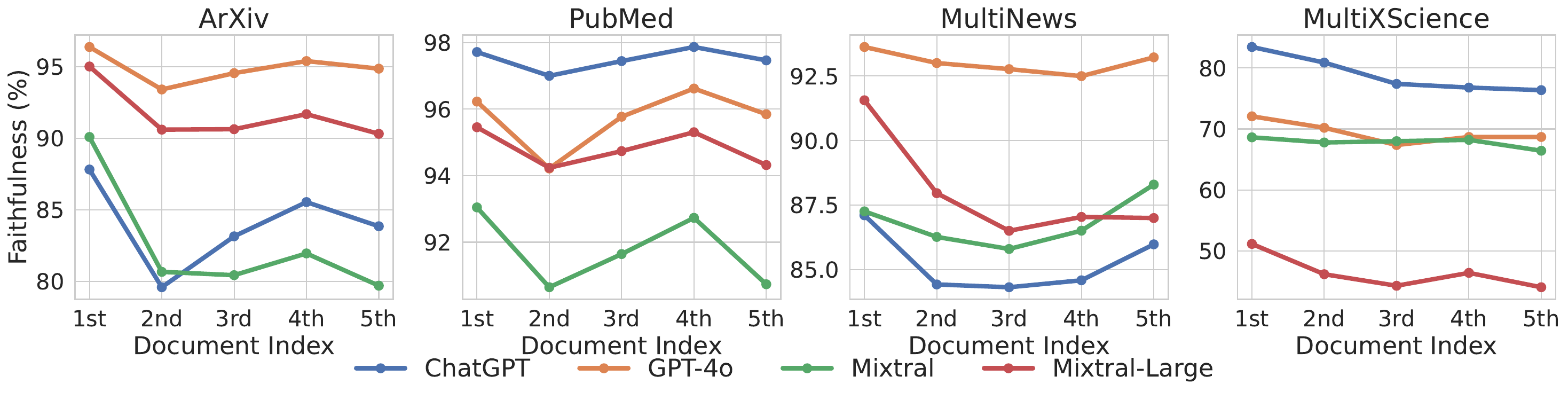}
    \caption{Faithfulness analysis across different positions using GPT-4o faithfulness metric.}
    \label{fig:faithfulness_gpt4o}
\end{figure*}
\subsection{Faithfulness Metric as Alignment and Attribution}\label{sec:superpal}
We evaluate whether the faithfulness score can be used for alignment. Specifically, we use SuperPal \cite{ernst-etal-2021-summary}, the state-of-the-art document-summary sentence alignment model, to test whether it reaches similar conclusions in terms of faithfulness. SuperPal operates at the sentence level, aligning summary sentences to document sentences. We use the indices of the aligned document sentences to compare alignment positions. The correlation between the two metrics is high, with a Spearman correlation of 0.74. We also verify the alignment visually, as shown in \autoref{fig:direct_analysis_superpal}. We observe the same trends as shown in \autoref{fig:direct_analysis} and discussed in Section~\ref{sec:faithfulness_analysis}: A dominant U-shaped curve for ArXiv, PubMed, and MultiNews, and a strong lead bias for MultiXScience. This demonstrates that using the maximum faithfulness score as an attribution method provides similar conclusions to those obtained from a trained document-summary sentence alignment model.

\subsection{Coverage Analysis} \label{sec:coverage_analysis}
We also provide the coverage analysis, by using the faithfulness score of all documents. We show the figure in \autoref{fig:coverage_analysis}. The trend generally follows the observation of \citet{ravaut-etal-2024-context}, who uses bigram matching between the summary and documents, observing either a U-shape or a lead bias.

\subsection{Faithfulness Analysis using GPT-4o}\label{sec:faithfulness_analysis_gpt4o}
Here, we now evaluate faithfulness using our best faithfulness metrics on a subset that excludes Llama-based models and includes only ArXiv, PubMed, MultiNews, and MultiXScience. Compared to \autoref{fig:faithfulness_gpt4o}, which uses MiniCheck, we observe a smoother graph, but the key findings remain the same. For example, we observe the same U-shape occurring for ArXiv, PubMed, and MultiNews, while MultiXScience exhibits a strong lead bias. As mentioned in Section~\ref{sec:faithfulness_experimental_setup}, since the computation is expensive, we still use MiniCheck for the remaining analyses.

\begin{table*}[!t]
    \centering
    \small
    \begin{tabularx}{\textwidth}{p{1.5cm} X}
        \toprule
        Method & Prompt \\
        \midrule
        Faithfulness evaluation & Document:\newline
[ARTICLE]\newline \newline
Sentence:\newline
[SUMMARY]\newline \newline
Determine if the sentence is factually consistent with the document provided above. A sentence is factually consistent if it can be entailed (either stated or implied) by the document. Please start your answer with “Yes.” or “No.” Please briefly explain the reason within 50 words.""" \\
    \midrule \midrule
    ArXiv generation & Read the following scientific paper. Produce a summary in 6 sentences. You must give your in a structured format: '''Summary: [your summary]''', where [your summary] is your generated summary.\newline
==========\newline
[ARTICLES]\newline
==========\\
\midrule
    PubMed generation & Read the following scientific paper. Produce a summary in 7 sentences. You must give your in a structured format: '''Summary: [your summary]''', where [your summary] is your generated summary.\newline
==========\newline
[ARTICLES]\newline
==========\\
\midrule
    MultiNews generation & Read the following news articles. Produce a summary in 10 sentences. You must give your in a structured format: '''Summary: [your summary]''', where [your summary] is your generated summary.\newline
==========\newline
[ARTICLES]\newline
==========\\
\midrule
    MultiXScience generation & Read the following abstracts. Produce a summary in 5 sentences. You must give your in a structured format: '''Summary: [your summary]''', where [your summary] is your generated summary.\newline
==========\newline
[ARTICLES]\newline
==========\\
    \midrule \midrule
    Focus prompt & [Generation Prompt]\newline
Pay special attention to the [top articles/articles in the middle/bottom articles].\\
\midrule
    Iterative prompt & Read the following section of a scientific paper.\newline
==========\newline
[NEXT DOCUMENT]\newline
==========\newline \newline
Below is a summary up until this point:\newline
==========\newline
[SUMMARY]\newline
==========\newline \newline
We are going over the articles sequentially to gradually update one comprehensive summary. Produce an updated summary in 6 sentences. You must give your in a structured format: '''Summary: [your summary]''', where [your summary] is your generated summary.\\
\midrule
Hierarchical merging & Below are several summaries:\newline
---\newline
[SUMMARIES]\newline
---\newline
Create one comprehensive summary by recursively merging summaries of its chunks. Despite this recursive merging process, you need to create a summary that seems as though it is written in one go. The summary must be within 6 sentences. You must give your in a structured format: '''Summary: [your summary]''', where [your summary] is your generated summary.\\
        \bottomrule
    \end{tabularx}
    \caption{Prompts for evaluation (top), standard generations (middle), and advanced generation techniques (bottom).}
    \label{fig:prompts}
\end{table*}
\section{Prompts}\label{sec:prompts}
We present the prompts used for evaluation and generation in Figure~\autoref{fig:prompts}. The faithfulness metric prompt is adapted from \cite{tang-etal-2024-tofueval}. For standard generation and focus prompt, we adapted the prompt from \cite{ravaut-etal-2024-context}. Finally, we adapt the iterative update and hierarchical merging prompts from \cite{chang2024booookscore}.

\end{document}